\documentclass[runningheads]{llncs}
\usepackage{graphicx}
\usepackage{subcaption}
\usepackage{amsmath,amssymb} % define this before the line numbering.
\usepackage{color}
\usepackage{algorithm}
\usepackage{algpseudocode}
\usepackage{ulem}
\usepackage{url}
\usepackage{cite}

\usepackage[width=122mm,left=12mm,paperwidth=146mm,height=193mm,top=12mm,paperheight=217mm]{geometry}
\usepackage[perpage]{footmisc}

\algnewcommand\algorithmicforeach{\textbf{for each}}
\algdef{S}[FOR]{ForEach}[1]{\algorithmicforeach\ #1\ \algorithmicdo}

\newcommand{\mysubsubsection}[1]{\vspace{0.2cm} \noindent
\underline{{\bf #1}} \vspace{0.1cm}}

\ifx\pdfoptionalwaysusepdfpagebox\relax\else
\fi

\begin{document}
% \renewcommand\thelinenumber{\color[rgb]{0.2,0.5,0.8}\normalfont\sffamily\scriptsize\arabic{linenumber}\color[rgb]{0,0,0}}
% \renewcommand\makeLineNumber {\hss\thelinenumber\ \hspace{6mm} \rlap{\hskip\textwidth\ \hspace{6.5mm}\thelinenumber}}
% \linenumbers
\pagestyle{headings}
\mainmatter
%\def\ECCV16SubNumber{1569}  % Insert your submission number here

%\title{Parallelizing Hypothesis Fusion for Rapid MAP inference}
\title{Multi-way Particle Swarm Fusion}

%\titlerunning{ECCV-16 submission ID \ECCV16SubNumber}

%\authorrunning{ECCV-16 submission ID \ECCV16SubNumber}

%\author{Anonymous ECCV submission}
%\institute{Paper ID \ECCV16SubNumber}

\author{Chen Liu$^*$\inst{1} \and Hang Yan$^*$\inst{1} \and Pushmeet Kohli\inst{2} \and Yasutaka Furukawa\inst{1}}
\institute{\{chenliu,yanhang,furukawa\}@wustl.edu
  \and pkohli@microsoft.com}

\maketitle

\begin{abstract}
  \renewcommand{\thefootnote}{\fnsymbol{footnote}}
  \footnotetext[1]{ indicates equal contribution.}
  \renewcommand{\thefootnote}{\arabic{footnote}}
This paper proposes a novel MAP inference framework for Markov Random
Field (MRF) in parallel computing environments.  The inference
framework, dubbed Swarm Fusion, is a natural generalization of the
Fusion Move method. Every thread (in a case of multi-threading
environments) maintains and updates a solution. At each iteration, a
thread can generate arbitrary number of solution proposals and take
arbitrary number of concurrent solutions from the other threads to
perform multi-way fusion in updating its solution. The framework is
general, making popular existing inference techniques such as
alpha-expansion, fusion move, parallel alpha-expansion, and hierarchical
fusion, its special cases. We have evaluated the effectiveness of our
approach against competing methods on three problems of varying
difficulties, in particular, the stereo, the optical flow, and the
layered depthmap estimation problems.
~\footnote{Project page: http://www.cse.wustl.edu/$\sim$chenliu/swarm-fusion.html}
%Our experiments have shown that the Swarm Fusion method outperform
%competing methods, especially for challenging problems with massive
%solution space.
% can effectively exploits the parallel computing resources and become
%effective especially for challenging problems.
%\yasu{must be at least 70 and at most 150 words.}
\end{abstract}

\section{Introduction}

Parallel computation has changed the field of computing.  In the 90s,
most processors had single cores. In 2016, processors have often 4 cores,
or even 8. Cluster computing further expands the potential
of parallel computation, where one can easily launch a processing job
using hundreds or even thousands of computational nodes in a cloud.
In the recent work on the AI program playing the ancient Chinese board
game of {\it Go}, parallelization plays a key role in the Monte-Carlo
tree search~\cite{silver2016mastering}.

Parallel computation offers tremendous potential for Computer Vision. As
image sensing technologies have gone through revolutions, we are in
ever growing demands in solving very large problems. One may need to
apply image denoising to 50 Megapixel images from latest digital
SLRs (e.g., Canon EOS 5DS), stitch thousands of images to generate gigapixel
panoramas~\cite{kopf2007capturing}, or solve volumetric reconstruction and
segmentation problems over a billion ($=1024^3$)
voxels~\cite{hane2013joint}.
%
%% However, state-of-the-art algorithms are still inherently
%% sequential in many Computer Vision problems. Take a Fusion Move
%% method
%% (FM)~\cite{fusion_flow,fusion_moves_for_markov_random_field_optimization,second_order_stereo}
%% for example, which has been one of the most effective techniques
%% for Markov Ran- dom Field (MRF) inference, including the above
%% problems like image denoising, image stitching, or volumetric
%% reconstruction. It sequentially improves solution by fusing the
%% current solution with a solution proposal. Its successful applica-
%% tions go beyond and also cover optical flow, stereo, image
%% inpainting, or image segmentation
%% problems~\cite{fusion_moves_for_markov_random_field_optimization} .
%
Markov Random Field (MRF) has been a very successful framework to solve
these problems in Computer Vision. However, state-of-the-art algorithms
for MRF inference are still inherently sequential. Take a Fusion Move
method
(FM)~\cite{fusion_flow,fusion_moves_for_markov_random_field_optimization,second_order_stereo}
for example, which has been one of the most effective techniques for MRF
inference. It sequentially improves solution by fusing the current
solution with a solution proposal. It has been successfully applied to
many problems such as optical flow, stereo, image inpainting, or image
segmentation~\cite{fusion_moves_for_markov_random_field_optimization}.

Unleashing the power of parallel computation for effective MRF inference
% Breaking the sequential nature of FM
would then bring fundamental contributions to Computer
Vision. Currently, FM suffers from a few vital limitations due to its
sequential nature. First, standard FM allows only two options per
variable in each fusion, either the current solution or a
proposal~\cite{fusion_moves_for_markov_random_field_optimization}. Second,
only a single proposal generation scheme is used in each fusion
step.~\footnote{Recently, an extension of FM was introduced for layered
depthmap estimation~\cite{layered_depthmap}, where a solution subspace, instead
of a single solution, is proposed and fused with the current
solution. However, this approach is also limited to the use of one proposal
generation scheme in each fusion.}
Our approach, dubbed {\it Swarm Fusion} method (SF), makes a few key
distinctions from existing approaches: 1) Multiple threads (or computing
nodes) simultaneously keep and improve solutions; and 2) Each fusion in
each thread can generate arbitrary number of solution proposals and use
arbitrary number of concurrent solutions in the other threads, to be fused
with the current solution.

We have evaluated the effectiveness of our approach over three problems
in Computer Vision, specifically, stereo, optical flow, and layered
depthmap estimation.
%
%Our approach outperforms all the other competing methods, especially for
%challenging problems with a large label space.
%
%Our idea is very simple. We believe that the method is easily
%reproducible with minimal coding
%
Our idea is extremely simple and the new inference framework can be
integrated into existing system with minimal coding. We believe that
this paper would have immediate impact on numerous Computer Vision
researchers or engineers, currently solving MRF problems with
conventional methods.

\section{Related work}

MRF inference has been a very active field in Computer Vision with
extensive literature. We refer the readers to survey articles for
comprehensive
reviews~\cite{middlebury_mrf,comparative_study_of_modern_inference}, and
here focus our description on closely related topics.

\mysubsubsection{Parallel Alpha-Expansion}

\noindent Lempitsky et
al.~\cite{fusion_moves_for_markov_random_field_optimization}
introduces parallel computation to the alpha-expansion technique,
where multiple threads simultaneously fuse mutually exclusive sets of
labels. Kumar et al.~\cite{hierarchical_graph_cuts_kumar_and_koller},
Delong et al.~\cite{delong_hierarchical_fusion}, and Veksler et
al.~\cite{olga_hierarchical_alpha_expansion} investigated hierarchical
approachs, where labels can be simultaneously fused from the bottom to
the top in a tree of labels.
Instead of taking a hierarchical approach, Batra et
al.~\cite{Dhruv_pushmeet_making_the_right_move} adaptively computed an
effective sequence of labels to explore. This technique can be combined
with parallel alpha-expansion techniques to obtain further speed-up.
Strictly speaking, these approaches are not in the family of Fusion Move
methods (FM), because they only consider constant label proposals. Our
approach is a generalization of FM.

\mysubsubsection{Parallel MAP inference}

\noindent The core MAP inference itself can be parallelized.  Strandmark
et al.~\cite{strandmark_parallel} parallelized graph-cuts.
Message passing algorithms are friendly to GPU implementation and can
exploit the power of parallel computation.
While state-of-the-art optimization libraries are often freely available
for non-commercial purposes, most companies have to develop and maintain
in-house implementation of these algorithms.  The core optimization
libraries are very complex and their modifications require significant
engineering investments. In contrast, our idea is extremely simple and
easily reproducible by standard engineers.

% \footnote{GPU speeds-up message-passing algorithms via parallel
% computation. However, these algorithms need to store all the messages
% and states and cannot handle problems with a large label
% space~\cite{layered_depthmap}.}

\mysubsubsection{Fusion Move methods}

\noindent FM was first introduced by Lempitsky et al.~\cite{fusion_flow}
in solving the optical flow problem. FM has been effectively used to
solve other challenging problems in Computer Vision such as stereo with
second order smoothness priors~\cite{second_order_stereo}, stereo with
parameteric surface fitting and segmentation (i.e. Surface
Stereo)~\cite{surface_stereo}, and multicut
partitioning~\cite{fusion_moves_for_correlation_clustering}.
FM has two main advantages over other general inference
techniques~\cite{TRW-S,loopy_belief_propagation}. First, FM allows us to
exploit domain-specific knowledge by customizing proposal generation
schemes. Second, FM can handle problems with very large label spaces
(and even real-values variables), because the core optimization
solves a sequence of binary decision problems.
In contrast, methods like message passing algorithms need to maintain
messages and beliefs for the entire label space all the time.
Although conceptually straightforward, we are not aware of {\it Parallel
  Fusion Move (PFM)} algorithms that fuse solution proposals, as opposed to
labels, in parallel. This paper seeks to fully unleash the power of
parallel computation based on FM in the most general setting.
% for the most general FM via a more general framework.

\mysubsubsection{Evolutionary algorithms and Particle Swarm Optimization}

\noindent
Genetic algorithms (GA)~\cite{ga} and Particle Swarm
Optimization (PSO)~\cite{pso} maintain multiple solutions and improve
them over time.
GA or PSO has been used to produce great empirical results,
e.g. in hand tracking~\cite{pushmeet_hand_tracking}.
At high level, our strategy is similar in spirit. However, GA or PSO
rather arbitrarily copies parts of the solutions or makes random movements
in each step (i.e., limited theoretical justification).
Our approach directly optimizes the objective function to improve
solutions.

\section{Multi-way particle swarm fusion} \label{section:algorithm}
Multi-way Particle Swarm Fusion is a natural extension of the Fusion
Move method (FM). We call our method Swarm Fusion (SF) in short. Let us
take multi-threading environment to explain our idea, while the
technique is also applicable to other parallel programming model such as
{\it MapReduce} in cloud computing.

Assuming we have $N$ threads $\{T_i | i=1, 2, \cdots, N\}$,
each thread $T_i$ maintains and updates a solution $S_i$ in
parallel.
%Alpha
%Expansion picks one label in each iteration to perform fusion. FM
%generates a solution proposal via a proposal generation scheme.
%invokes one proposal generation scheme and generates a solution proposal
%for fusion.
SF has 1) a {\it proposal generator} for each thread which picks
arbitrary number of proposal generation schemes and generates proposals,
and 2) a {\it solution pool}, from which a thread picks arbitrary number
of intermediate solutions generated by the others.
% threads.
%
% Traditional FM has a set of proposal generation schemes to be
% invoked
%
% There are two places to
% collect solutions or solution proposals for fusion. First, the {\it
% proposal pool} can generate new solution proposals on request.
% %
% Second, the {\it solution pool} stores the solutions that have been
% fused by the threads.
In our base configuration, the solution pool remembers $N$ best
solutions, one from each thread.

SF has two main parameters $\alpha_i$, $\beta_i$ (for each thread
$T_i$), determining its behaviors: In each fusion step, a thread
generates $\alpha_i$ solution proposals using its proposal generator, and collects
% \chen{A proposal pool sounds confusing. We often don't
% really maintain such a pool as many proposals are
% generated dynamically based on current solution. We just
% have certain strategies right there and in each iteration,
% we pick some strategies to generate proposals on the
% fly. This is different with the solution pool which is
% really there.}
$\beta_i$ solutions from the solution pool, based on a user-defined
strategy or at random to be simple. The values of $\alpha_i$ and
$\beta_i$ can vary per iteration for flexibility.
The thread then fuse all these proposals and/or solutions to find a solution
with lower energy state and update the solution pool accordingly.
%
% One can further customize how to pick certain proposal generation
% schemes (as in ~\cite{delong}) or solutions.

\begin{figure}[tb]
 \includegraphics[width=\columnwidth]{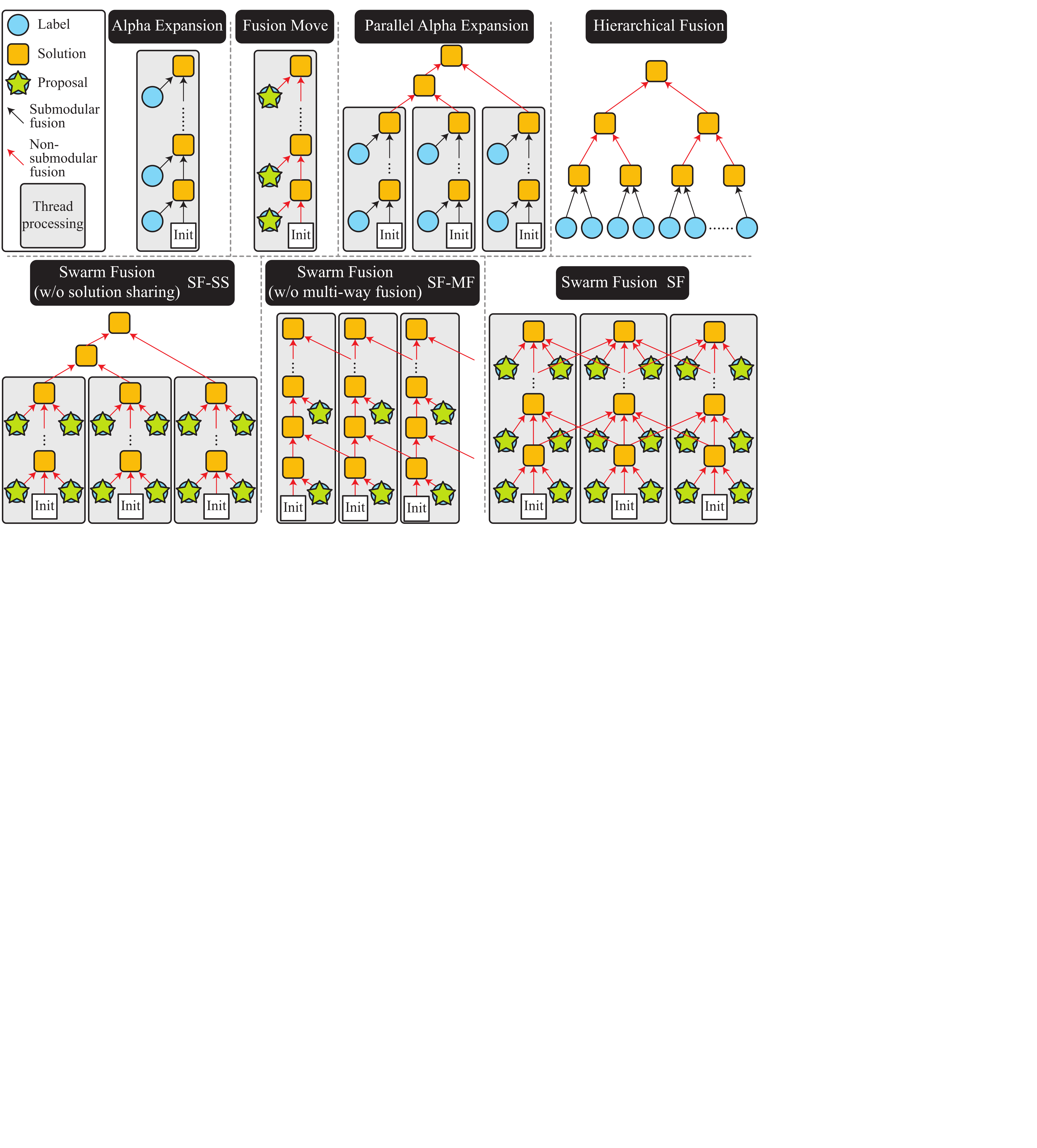} \caption{Swarm
   Fusion (SF) architecture and its relationships to existing methods. The
 bottom right example shows the general SF architecture, where
 each thread takes arbitrary number of solution proposals and concurrent
 solutions for fusion. The framework is flexible and
 can realize other data processing architectures depending on the
 parameters (e.g., the left two examples in the bottom row).
 %
 % The left two examples at the
 % bottom row are other Swarm fusion architectures with some restrictions,
 % for which we provide comparative evaluations in
 % Section~\ref{section:results}.
 %
% The left two examples at the bottom are other Swarm fusion
% architectures with some restrictions, to be used for evaluations later.
 %
 It is easy to verify that existing popular MRF inference methods such
 as Alpha Expansion~\cite{alpha_expansion}, Fusion Move~\cite{fusion_moves_for_markov_random_field_optimization},
 Parallel Alpha Expansion~\cite{olga_hierarchical_alpha_expansion}, and Hierarchical
 Fusion~\cite{delong_hierarchical_fusion}, are all special cases of
 SF.}
\label{fig:model}
\end{figure}

Swarm Fusion framework is very flexible and yields various data
processing architectures as shown in Fig.~\ref{fig:model}.
The bottom right architecture is the most general one, in which
threads conduct multi-way fusion of their current solution, proposals
from their own proposal generators and/or concurrent solutions.
For general non-submodular energy, we use TRW-S~\cite{TRW-S} for inference.
%% And the energy is not submodular
%% in general, for which effective inference techniques such as
%% TRW-S~\cite{TRW-S,opengm} exist.
%
%, and hence we use TRW~\cite{kolmogorov} for the fusion, while any other
%compatible inference techniques can be used~\cite{opengm}.
However, if one knows that a certain fusion step is a binary fusion with
submodular energy, one can use
alpha-expansion~\cite{alpha_expansion}.
%
%If a certain fusion step is a
%binary fusion with non-submodular energy, one can use
%QPBO~\cite{second_order_stereo}.
%
QPBO~\cite{QPBO} can be used to perform binary fusion with non-submodular energy.
Note that the threads appear synchronized in the figure only for
illustration purpose. In practice, all the threads run asynchronously
with a (read-write) lock on the data in the solution pool (See
Algorithm~\ref{algorithm:sf}).
%The asynchronous computing further unleashes the power of parallism.
%The asynchronous processing could be difficult for cloud parallel
%computing.
%
%
%
\begin{algorithm}[t]
 \caption{Swarm Fusion method}
 \label{algorithm:sf}
 \begin{algorithmic}
  %\Procedure{Swarm Fusion method} {}
  \Procedure{} {$\alpha, \beta$}
  \State $\mathcal{S}_{pool} \leftarrow \emptyset$ \ \ \ //Solution pool
  %\State Initialize $\mathcal{P}_{pool}$ // Proposal pool
  \ForEach{thread $T_i$}
  \State Initialize its solution $S_i$
  \EndFor
  \State
  \ForEach{thread $T_i$ in parallel till convergence}
  \State Generate $\alpha_i$ solution proposals $\mathcal{P}$ %\subset \mathcal{P_{\mbox{pool}}}$
  \State Pick $\beta_i$ solutions $\mathcal{S} \subset \mathcal{S_{\mbox{pool}}}$
  \State $S_i \leftarrow \mbox{Fuse}(S_i, \mathcal{P}, \mathcal{S})$
  \State Replaces the solution in $\mathcal{S}_{pool}$ with $S_i$
  %\State [ Generates proposals and update $\mathcal{P}_{pool}$, if necessary ]
  \EndFor
  \EndProcedure
 \end{algorithmic}
\end{algorithm}

\mysubsubsection{Relationships to existing methods}

\noindent It is easy to verify that Alpha-Expansion
(AE)~\cite{alpha_expansion}, Fusion Move
(FM)~\cite{fusion_moves_for_markov_random_field_optimization},
Parallel Alpha Expansion
(PAE)~\cite{fusion_moves_for_markov_random_field_optimization},
and Hierarchical Fusion
(HF)~\cite{delong_hierarchical_fusion,olga_hierarchical_alpha_expansion}
are all special cases of the Swarm Fusion method (SF). AE
can be realized by setting $(\alpha=1, \beta=0)$ and
restricting the proposals to be constant labels with
a single thread. The same goes for FM, this time, without
the restriction on the proposal generation scheme. PAE is
realized by setting $(\alpha=1,\beta=0)$ with multiple
threads, again with a restriction on the proposal generation
scheme (the last sequential fusion in PAE is realized by
$(\alpha=0, \beta=1)$ with a single thread).
%
% The last sequential fusion (at the top of the example in the figure) can
% be modeled by changing to $\alpha=0, \beta=1$ with a single thread.
HF has a slightly different data processing model, without strong ties
between threads and data, but can be realized by setting ($\alpha=2,
\beta=0$) at the bottom level and ($\alpha=0, \beta=2$) at the remaining
levels, while allowing $S_i$ not to be used in the fusion steps of
$T_i$.

\section{Swarm Fusion instantiation}
% \hang{I didn't understand the organization of this section at the
%   first glance. Should it be clearer if we put 'Swarm fusion
%   architectures' in front of 'Competing methods' in each problem?}

\begin{figure}[tb]
  \includegraphics[width=\columnwidth]{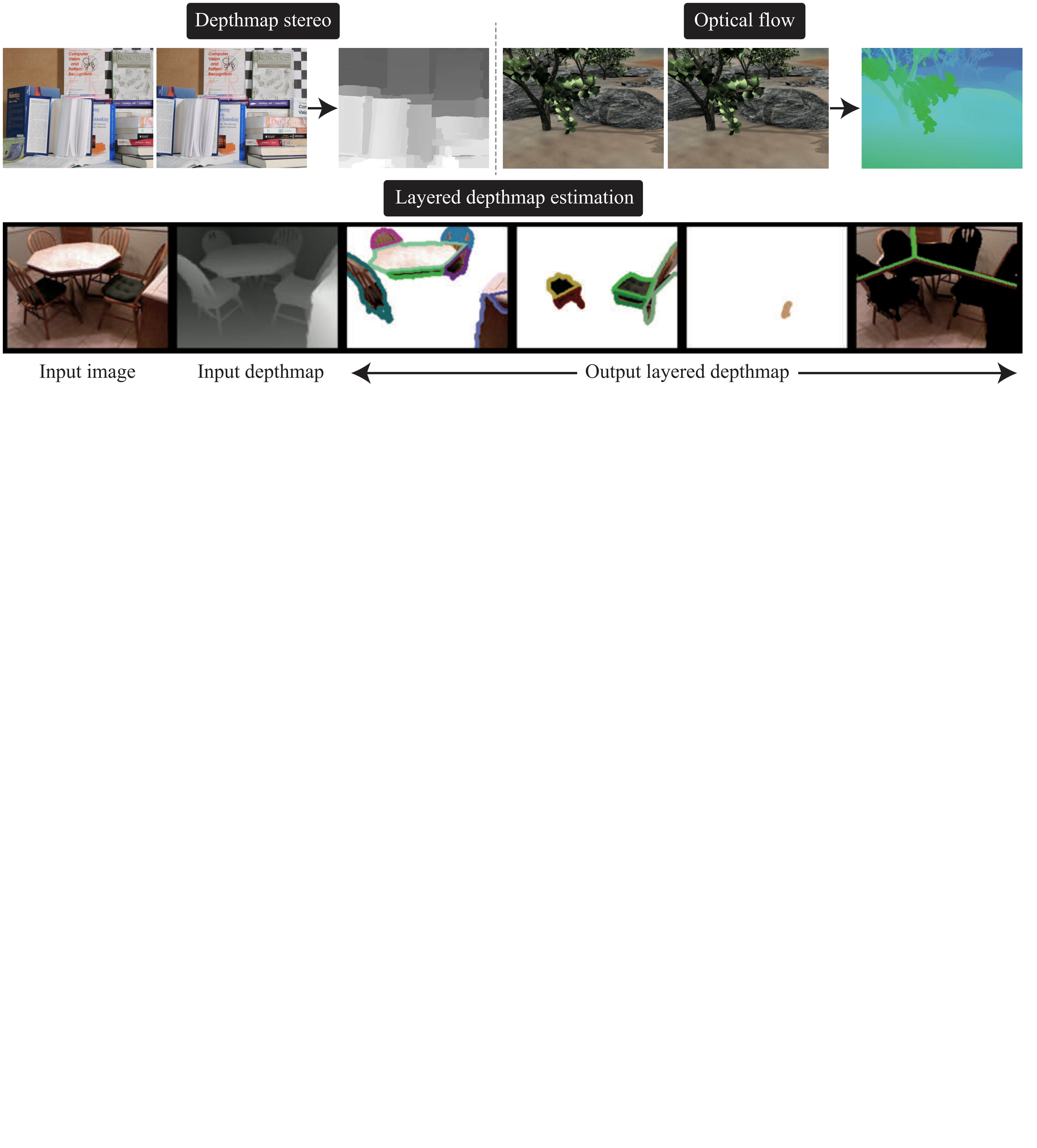} \caption{We
    compare our Swarm Fusion method against competing approaches on the
    depthmap stereo~\cite{middlebury_stereo}, the optical
    flow~\cite{middlebury_optical_flow} and the layered depthmap
    estimation~\cite{layered_depthmap} problem. In the layered
    depthmap problem, the input is a RGBD image, and the output is
    multiple layers of depthmaps. Each layer is a piecewise smooth
    parametric surface model.}\label{fig:problem}
\end{figure}

We compare SF against competing approaches over three problems in
Computer Vision, specifically, stereo, optical flow, and layered
depthmap estimation (see Fig.~\ref{fig:problem}).

% SF framework has high degrees of freedom and the
%challenge is to properly quantify the contributions of various
%algorithmic aspects.

%For this purpose, we have used the three SF architectures illustrated in
%Fig.~\ref{fig:model}.
%% For some problems, we have further limited the capabilities of SF
%% on purpose, to enable effective and fair comparative
%% evaluations. We now explain the details of the three problems.

% We pick three representative problems in Computer Vision to evaluate the
% effectiveness of the Fusion swarm methods against existing methods. We
% control the SF architectures for each problem independently to make
% compa
%
%\yasu{this section is very rough at the moment. requires polishing. but
%algorithms might change as we run experiments.}  We now explain detailed
%implementation of the Swarm Fusion method for three problems in Computer
%Vision (See Fig.~\ref{fig:problem}).  5
\subsection{Swarm Fusion stereo}
We start with a simple depthmap stereo problem with standard unary and
pairwise terms. We employ submodular pairwise terms to make this
stereo represent relatively ``easy'' MRF inference problem.
The unary terms are computed as the average robust photoconsistancy
score~\cite{second_order_stereo} between the reference image and the others
inside a $7\times 7$ pixels window. The pairwise terms are simple
truncated absolute label difference with maximum label difference
$\sigma_s=4$. The total energy is defined by the sum of the two, while
scaling the pairwise terms by a factor of $0.005$. For simplicity we
do not enforce the visibility constraint.

\mysubsubsection{Competing methods}

\noindent For simple stereo problems with submodular energy as ours, the
sophistication of photometric consistency
function~\cite{mvs_furukawa_survey} makes unary terms highly
informative, where efficient inference algorithms such as graph-cuts
exist.
%Our experiments have also supported this, where the fusion method
%(i.e., the use of proposal generation) rather makes it slow due to the
%overhead, while not improving final energy.
Therefore, we have chosen algorithms based on Alpha-Expansion, namely
single thread Alpha Expansion(AE), Parallel Alpha
Expansion(PAE)~\cite{fusion_moves_for_markov_random_field_optimization}
and Hierarchical Fusion(HF)~\cite{olga_hierarchical_alpha_expansion} to
be competing methods. For HF, we use Alpha-Expansion at the leaf node of
the label tree and QPBO in the other cases.
%when either of the
%child node is leaf node (constant label) and QPBO for all other nodes.

\mysubsubsection{Swarm Fusion architectures}

\noindent The three swarm architectures in Fig.~\ref{fig:model} have
been evaluated: SF-MF (SF without multi-way fusion), SF-SS (SF without
solution sharing), and the standard SF.
SF-MF implies $\alpha+\beta=1$, where each thread repeats
fusing a solution proposal ($\alpha=1, \beta=0$) for four iterations
by Graph-cuts and fusing a concurrent solution  ($\alpha=0, \beta=1$)
for one iteration by QPBO.
%expand 4 labels (repeat $\alpha=1, \beta=0$ for 4 iterations) with Alpha
%Expansion and then fuse one solution from other threads ($\alpha=0,
%\beta=1$) by QPBO.
In the later case, a thread randomly chooses one solution from the
solution pool for fusion. SF-SS implies $\beta=0$, where $\alpha$ is
the free parameter and set to 4. In this case one thread will fuse 4
labels, together with current solution in that thread by TRW-S in each
iteration and never exchanges solutions with other threads. We perform
a multi-way fusion of solutions from all the threads at the end to
obtain a final solution
% at
% the end to fuse solutions from all the threads to a single solution
(similar to PAE). For standard SF architecture, we have used
($\alpha=4, \beta=1$).  To make the comparison simple, we restrict our
solution proposals to be constant-label proposals.
%
% Due to the lack of multi-way fusion, the swarm fusion architecture for
% the stereo problem is represented by the middle example at the bottom of
% Fig.~\ref{fig:model}.

\subsection{Swarm Fusion optical flow}
\label{section:optical_flow}
Fusion Move was first introduced by Lempitsky et al.~\cite{fusion_flow} to
solve the optical flow problem.
% Optical flow was the problem, in which the Fusion Move method was first
%introduced by Lempitsky et al.~\cite{viktor}.
We copy their problem setting and use images from the Middlebury optical
flow benchmark~\cite{middlebury_optical_flow}. We share similar proposal
generation schemes with Lempitsky et al~\cite{fusion_flow}
%instead of the HS algorithm, we
 with some modifications.~\footnote{First, we use more recent
   Farneback algorithm and change the level of pyramids from 1 to 5,
   then use either 3, 5 or 7 for parameter ``polyN''. Besides the
   clustering idea, we add three simple proposal generation schemes
   based on the current solution as suggested in
   \cite{fusion_flow}. In \textit{shift proposal}, the flow field in
   the current solution is shifted in either x or y directions for
   either 1, 2 or 3 pixels. In \textit{stagger proposal}, the flow
   field is shifted by a vector randomly drawn from a Gaussian
   distribution. In \textit{perturb proposal}, each flow value in the
   field is independently shifted by a vector randomly drawn from a
   Gaussian distribution. We choose schemes randomly when generating
   proposals.}

\mysubsubsection{Competing methods}

\noindent Fusion Move method in Lempitsky's paper is the first natural
contender. While they did not consider parallel implementation, it is
straightforward to combine the idea of Parallel Alpha Expansion and
Fusion Move. Therefore, the second competing method is ``Parallel Fusion
Move''(PFM), which is equivalent to Parallel Alpha Expansion with constant
label solutions replaced by solution proposals.
%each thread repeats generating
%a solution proposal and fusing it to the current solution. Solutions
%from all the results are sequentially fused to obtain the final solution
%as in the Parallel Alpha-Expansion method.
%
One problem of PFM is that infinite number of solution proposals can be
generated in their algorithm, and we do not know when to stop and
perform the final sequential fusion (See Parallel Alpha Expansion
architecture in Fig.~\ref{fig:model}). In our experiments, we manually
picked time limits to initiate the final fusion to make the comparisons
fair.
%In Fig.~\ref{fig:model}, this computational framework corresponds to
%``parallel alpha-expansion'' with the label solutions replaced by
%solution proposals.
%
The last contender is the mix of the Hierarchical Fusion and the Fusion
Move methods, dubbed ``Hierarchical Fusion Move''(HFM), where they
start from solution proposals as opposed to constant labels. One problem
of HFM is that we need to generate all the proposals first to build the
fusion tree. This undermines the power of fusion move that can
adaptively generate proposals based on the current solution. In our
experiments, we have manually generate 250 proposals at the
beginning.
%We first generate 25 proposals using LK and Farneback algorithms which
%are independent and generate other proposals based on these initial
%proposals).
The fusions are binary in these methods and we have used QPBO.

\mysubsubsection{Swarm Fusion architectures}

\noindent
%The three swarm fusion models (at the bottom of Fig.~\ref{fig:model})
%are used to evaluate the effectiveness of our method.
The three swarm architectures in Fig.~\ref{fig:model} (SF-MF, SF-SS, SF)
have been evaluated against the competing methods. For SF-MF, each
thread repeats generating solution proposals ($\alpha=1, \beta=0$) for
four iterations and fuses with one solution from others ($\alpha=0,
\beta=1$) for one iteration. This pattern is repeated. For SF-SS, each
thread generates three solution proposals for fusion in each iteration
($\alpha=3$, $\beta=0$). For SF, we repeat four iterations of
($\alpha=3, \beta=0$) and  one iteration of ($\alpha=0, \beta=3$).
%%
%Note that we have used the same set of proposal generation schemes as
%the competing methods.
We have used TRW-S for multi-way fusion and QPBO for binary fusion.
% , while our fusions are multi-way, for which we have used TRW-S.
%
% The full swarm fusion model (at the bottom right in
% Fig.~\ref{fig:model}) will be used to evaluate the effectiveness of our
% method. We will also try the other two Swarm fusion variants in
% Fig.~\ref{fig:model} with restrictions.
%
% 2 k-way 1 solution sharing. \chen{?}

\subsection{Swarm Fusion layered depthmap estimation}
Our last problem is layered depthmap estimation, recently proposed
in~\cite{layered_depthmap} (see the anonymous paper in the
supplementary material).  The problem seeks to infer layered depthmap
representation from a RGBD image, where each layer is a piecewise
smooth segmented depthmap. This is essentially a multi-layer extension
of Surface Stereo algorithm~\cite{surface_stereo}.
Layered depthmap estimation is a very challenging MRF inference problem
due to its massive solution space. The number of labels per pixel is
exponential in the number of layers, and is usually between 100,000 and
10,000,000. We copy their problem formulation and the proposal
generation schemes.~\footnote{Authors have proposed a novel fusion
scheme, where a solution subspace instead of a single solution is
generated by a proposal generation scheme. Since a solution subspace can
be represented by a concatenation of multiple solution proposals,
% \chen{It is a good explanation for this paper's purpose, but from the perspective of our previous work, I am reluctant to say a subspace equals to multiple solution proposals. When we design proposals, we find the subspace directly and there's no concept of solution proposals (we later simplify our subspace proposals to binary labels just for the comparison reason). It is essentially the same, but a subspace proposal is conceptually more general than concatenated solution proposals.},
their algorithm can be easily integrated into our Swarm Fusion framework.
However, competing fusion methods (e.g., Parallel Fusion Move or
Hierarchical Fusion) cannot handle a solution subspace proposal, making
it impossible to conduct fair comparative evaluations. We choose to use
a simple solution proposal for this experiment.}
%Lastly, we will evaluate Swarm fusion on a very challenging layered
%depthmap estimation problem~\cite{layered_depthmap}, which has been
%recently proposed (see the anonymized pdf in the supplementary
%material).

\mysubsubsection{Competing methods}

\noindent In this problem setting, solution proposals depend heavily on the current solution,
eliminating the possibility of using Hierarchical Fusion Move (HFM), which needs to
enumerate all the proposals to start. Therefore, viable competing
methods are Fusion Move (FM) and Parallel Fusion Move (PFM) as in the optical flow problem.
%
%We will also evaluate the standard Fusion Move method.
%
The fusions are binary, for which we use QPBO.

%The number of labels is extremely large (an order of 10,000 \yasu{what
%were the number of labels in our cvpr16?}), which eliminates the
%possibility of using hierarchical fusion. Therefore, we will evaluate
%fusion move and parallel fusion move algorithm as in the case of optical
%flow problem, where one solution proposal repeats being fused via QPBO.

\mysubsubsection{Swarm Fusion architectures}

\noindent The three swarm architectures with the same configurations as
in the optical flow problem have been evaluated.
%in Fig.~\ref{fig:model} are evaluated.
%The architecture parameters are exactly the same as in the optical flow problem:
%$\alpha=2$ ($\beta=0$) in SFWSS, the iteration of
%($\alpha=1, \beta=0$) and ($\alpha=0, \beta=1$) in SFWMF, and
%($\alpha=1, \beta=1$) for the standard SF.
%
%The fusions are multi-way, for which we use TRW-S.

\section{Experimental results}
\label{section:results} We have implemented the algorithms with
multi-threading support from C++ 11, and conducted the experiments on
Linux PCs with Intel Core i7 4790 processor with 4 cores. We have used
the Graph-cuts optimization code written by Veksler, using the libraries
provided by Boykov and
Kolmogorov~\cite{middlebury_mrf,alpha_expansion,what_energy_can_be_min_by_gc,mrf_experimental}.
% [2] Fast Approximate Energy Minimization via Graph Cuts.
%         Y. Boykov, O. Veksler, and R. Zabih.
%         In IEEE Transactions on Pattern Analysis and Machine Intelligence
%         (PAMI), vol. 23, no. 11, pages 1222-1239, November 2001.  
%
%     [3] What Energy Functions can be Minimized via Graph Cuts?
%         V. Kolmogorov and R. Zabih. 
%         In IEEE Transactions on Pattern Analysis and Machine Intelligence
%         (PAMI), vol. 26, no. 2, pages 147-159, February 2004. 
%         An earlier version appeared in European Conference on Computer
%         Vision (ECCV), May 2002.
%
%     [4] An Experimental Comparison of Min-Cut/Max-Flow Algorithms for
%         Energy Minimization in Vision. 
%         Y. Boykov and Vladimir Kolmogorov.
%         In IEEE Transactions on Pattern Analysis and Machine Intelligence
%         (PAMI), vol. 26, no. 9, pages 1124-1137, September 2004. 
We have used the QPBO and TRW-S implementations by
Kolmogorov~\cite{QPBO, TRW-S}. We have used 4 threads
for experiments unless indicated. We now look at our experimental
results for the three problems.

% To evaluate the effectness of different parallel structures, we plot
% the energy minimization process against time. We define the energy of
% a parallel optimization system at a certain time as the minimum energy
% of all threads at that time.

\mysubsubsection{Stereo}

\noindent We have chosen 7 images with the resolution of $695\times555$
from the Book sequence of Middlebury stereo
dataset~\cite{middlebury_stereo}. The number of disparity labels is set
to 256.
Since the order of labels is important for the expansion techniques,
we have used the same random order for all algorithms to avoid any
bias. 
%
% In PAE, SF-MF, SF-SS and SF algorithms, the proposal generator in each
% thread can only generate constant-label proposals within a evenly split
% subset of all labels.

\begin{figure}[tb]
\centering
\includegraphics[width=0.8\textwidth]{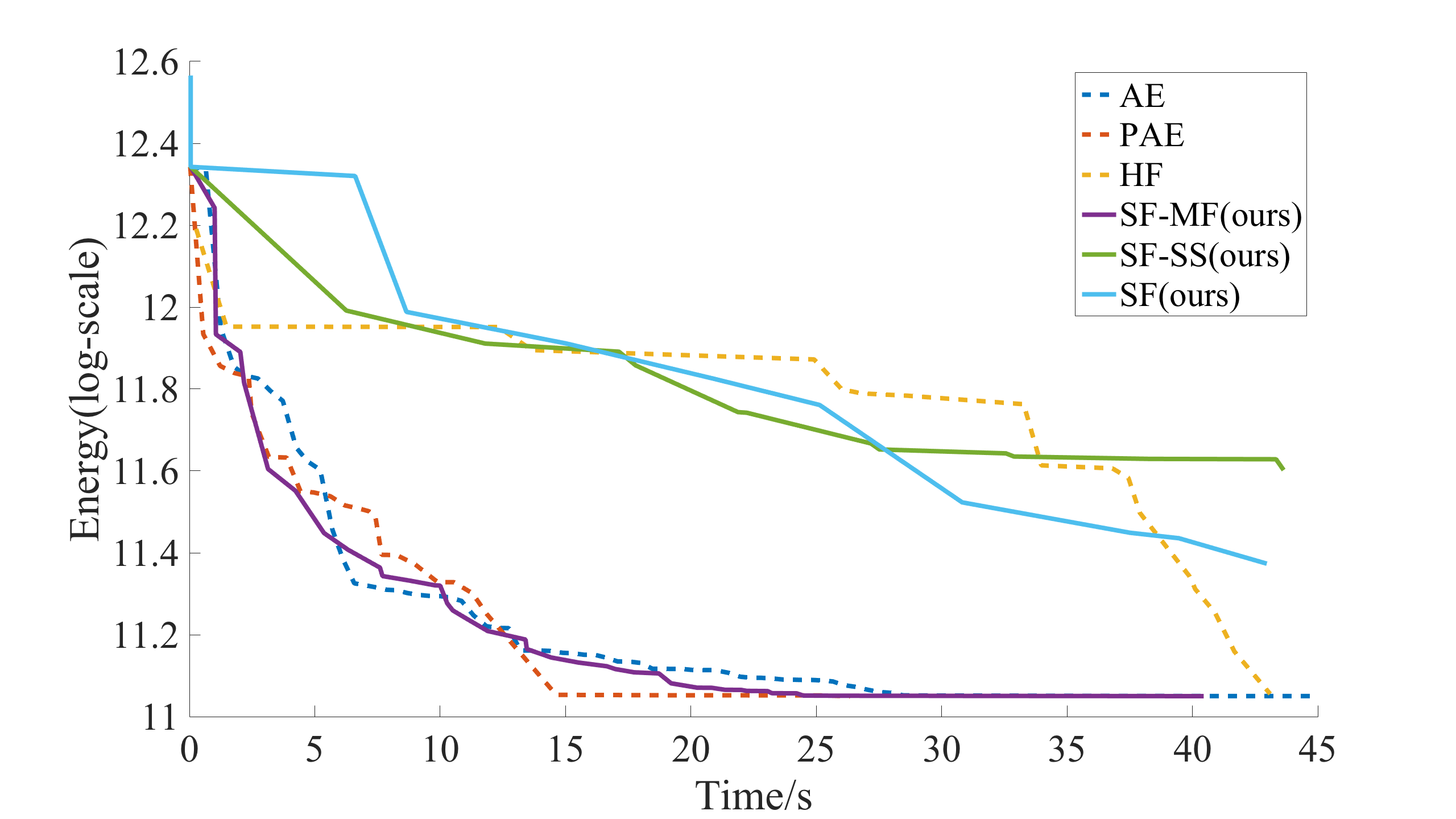}
\caption{Energy plots for the stereo problem. For a multi-threading algorithm, the energy at a given moment is
defined to be the best solution so far.}
\label{fig:stereo_global}
\end{figure}

\begin{figure}[!h]
\centering
\begin{subfigure}{0.49\textwidth}
\includegraphics[width=\textwidth]{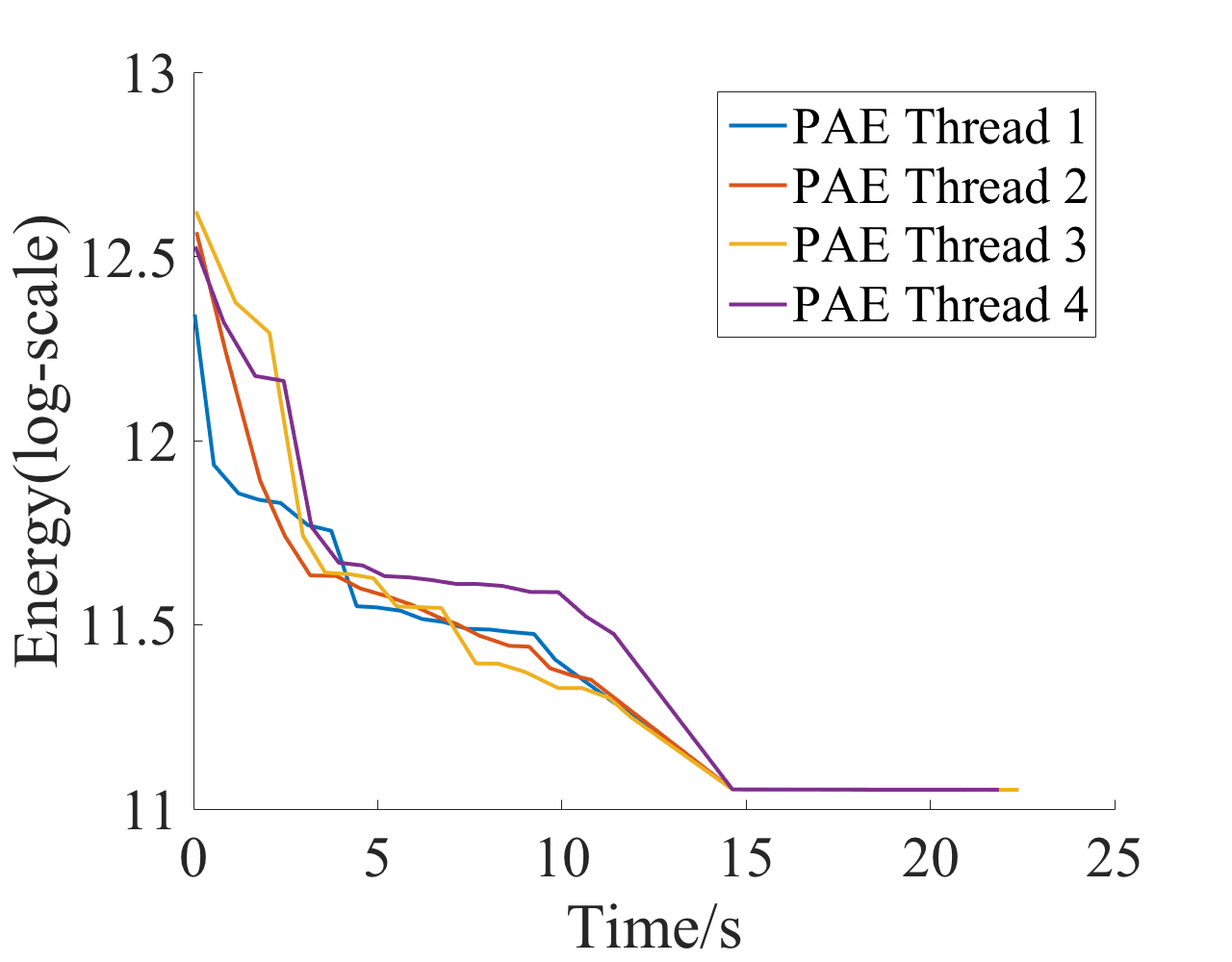}
\end{subfigure}
\begin{subfigure}{0.49\textwidth}
\includegraphics[width=\textwidth]{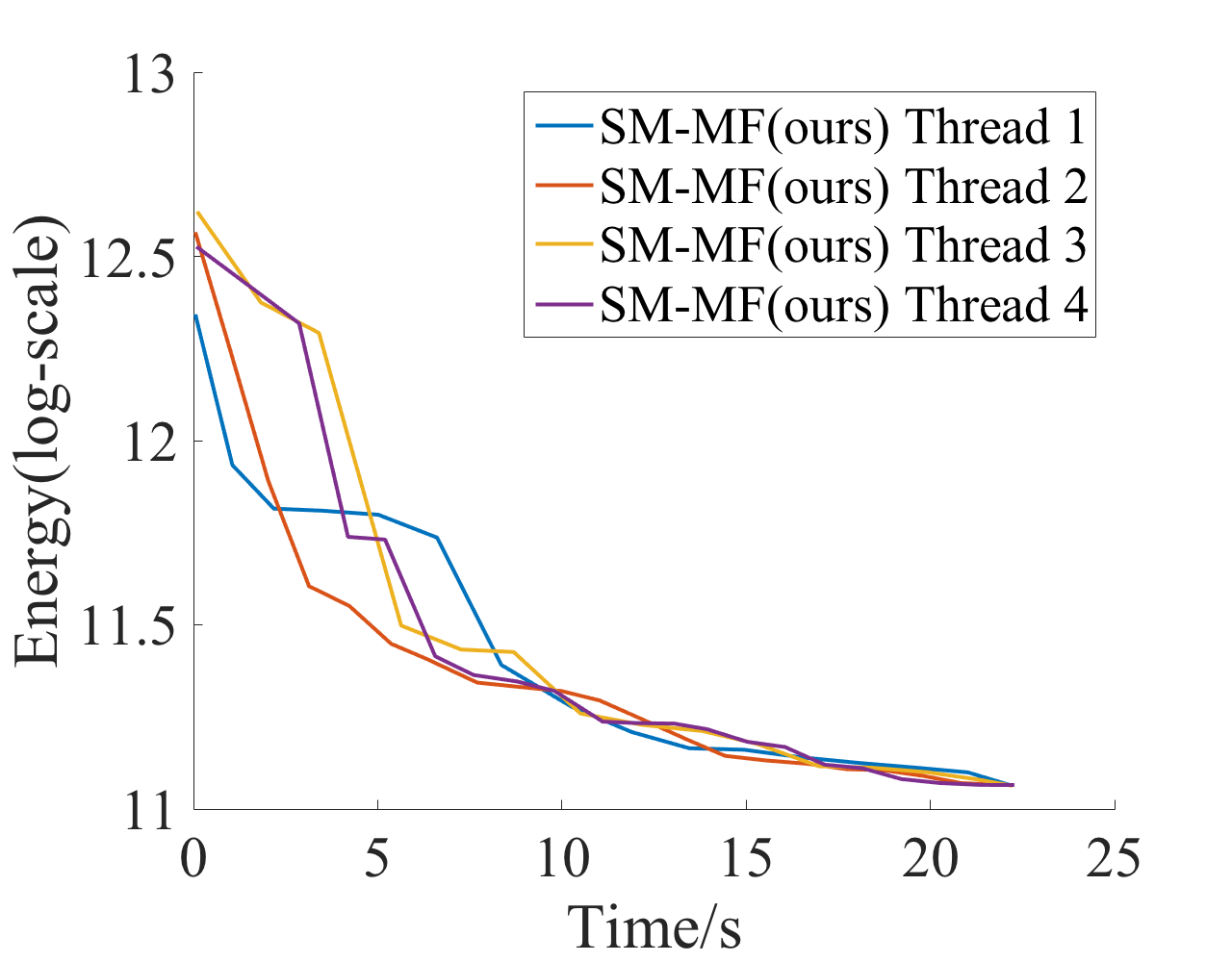}
\end{subfigure}
\caption{Energy plots per thread for the stereo problem. Left: PAE. Right: SF-MF(ours).}
\label{fig:stereo_threads}
\end{figure}

Figure~\ref{fig:stereo_global} compares the converegence rate of the
competing methods. Note that we define the energy of a multi-threading
system to be the energy of the best solution found so far.
%
%One unusual outcome is that Sequential Alpha Expansion and Parallel
%Alpha Expansion (PAE) have the same convergence speed, indicating that
%the stereo problem is a very easy one.
PAE and SF-MF converge faster than the single thread AE. However, the
speedup is not significant, which confirms the fact that the problem
is an easy one.
%This is well known in a community and this fact was rather expected.
Our approaches with multi-way fusion (SF-SS or SF) are the slowest
kind, because the TRW-S for multi-way fusion is slower than multiple
Alpha Expansion steps, and this stereo problem is too easy to gain
benefits through mulit-way fusion.
%HF does not perform well because the algorithm has to perform QPBO for
%all nodes with two non-leaf children.
%
%Another interesting observation is that the Hierarchical Fusion can
%achieve good energy only when they merge solutions at the top of the
%tree.
%
%  both PAE and SF-MF converge faster
% than sequential method. However, since fusing solutions with multiple
% labels by QPBO is slower than a single $\alpha$-expansion, PAE method
%
% Finally, the line of hierarchy fusion only makes
% jump when fusing on root node of the tree. This is because any fusion
% step on non-root nodes only have partial label information.
%We recoreded both single thread energy and system energy against
%time.
%
Figure~\ref{fig:stereo_threads} shows the energy plot per thread for
SF-MF (ours) and PAE. With solution sharing, the energy in SF-MF
decreases more uniformly, while in PAE the energy makes dramatic
decrease at the final fusion.
%
%The graph for PAE again confirms that the
%stereo is an easy problem, because all the threads are quickly
%converging to a solution before the final sequential fusion.
%shows the per-thread energy
%minimization process in SF-MF and PAE. The per-thread energy in SF-MF
%architecture decreases more uniformly than that of PAE. In the
%scenario where we need to query the best solution so far before the
%whole optimization converges, SF-MF architecture is a better choice.

For an easy optimization problem such as stereo with strong unary
terms and submodular pairwise terms, our full architecture with
solution sharing and multi-way fusion actually makes convergence
slower compared with PAE due to its overhead.
% of sophisticated fusion.

% over-sophisticated fusion algorihtm and multi-threading
% overhead. However, we can easily configure the architecture to make it
% better fit the problem, e.g. turn off multiway fusion and/or solution
% sharing.

% We define the energy of
% a parallel optimization system at a certain time as the minimum energy
% of all threads at that time.

\mysubsubsection{Optical Flow}

\begin{figure}[!ht]
  \centering
  \includegraphics[width=0.8\columnwidth]{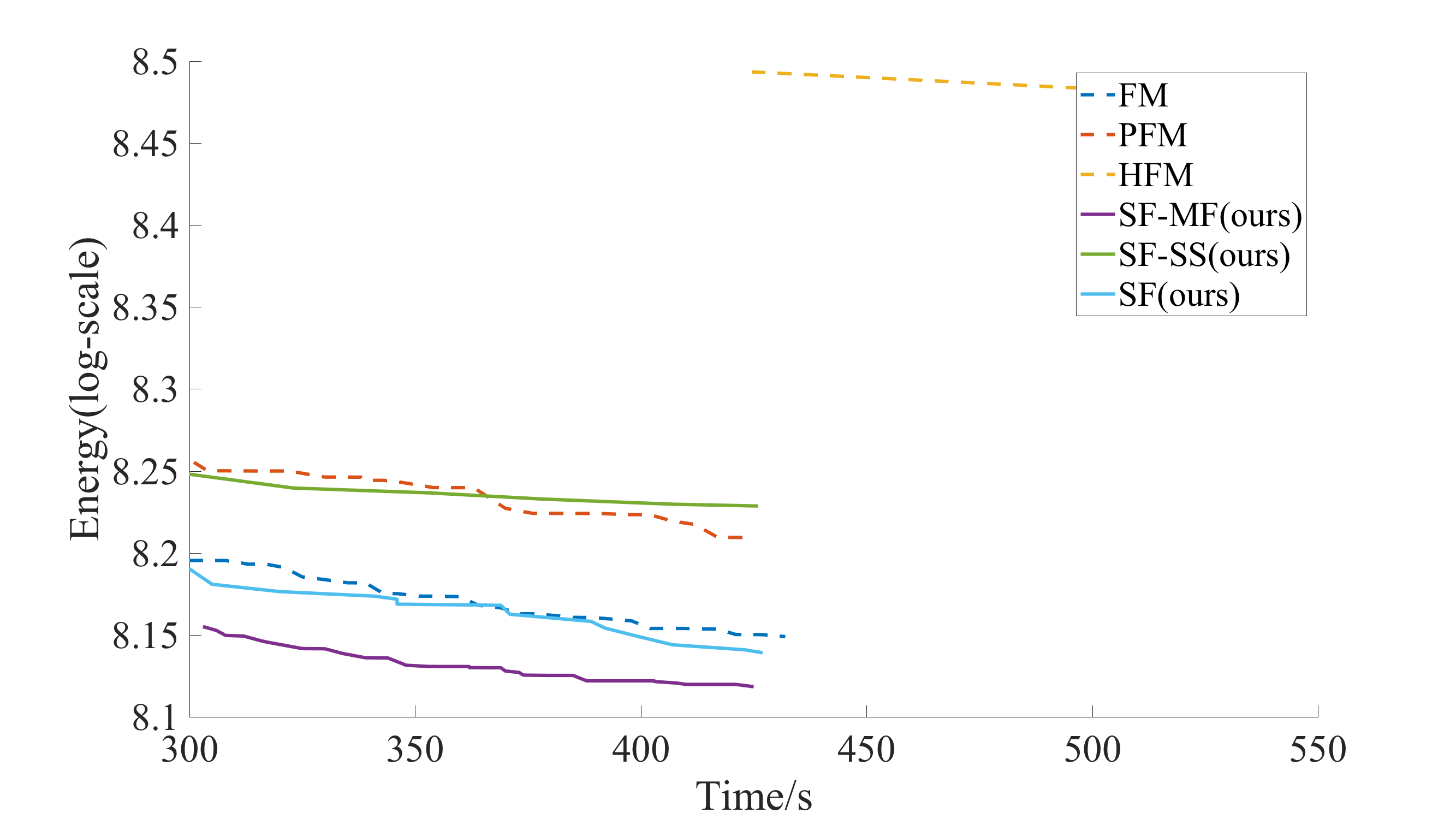}
  \caption{Energy plots for the optical flow problem. SF-MF has the best
    performance due to its solution sharing strategy.}\label{fig:optical_flow_convergence}
\end{figure}

\noindent
We have chosen the Dimetrodon image pair from the Middlebury flow
dataset~\cite{middlebury_optical_flow}. Figure~\ref{fig:optical_flow_convergence}
shows the energy plots of the three competing methods, Fusion Move
(FM), Parallel Fusion Move (PFM), and Hierarchical Fusion Move (HFM),
against our Swarm Fusion methods (SF-MF, SF-SS, SF). A key observation
is that SF-MF converges quicker and better than PFM. This is indeed the
benefits of solution sharing in our network. Optical flow is a more
difficult problem and many solution proposals are not effective.
The solution sharing (i.e., SF-MF) allows all the threads to exchange
effective solution proposals in the middle of the
optimization.

\begin{figure}[!ht]
  \centering
  \begin{subfigure}[b]{0.49\columnwidth}
    \centering
    \includegraphics[width=\columnwidth]{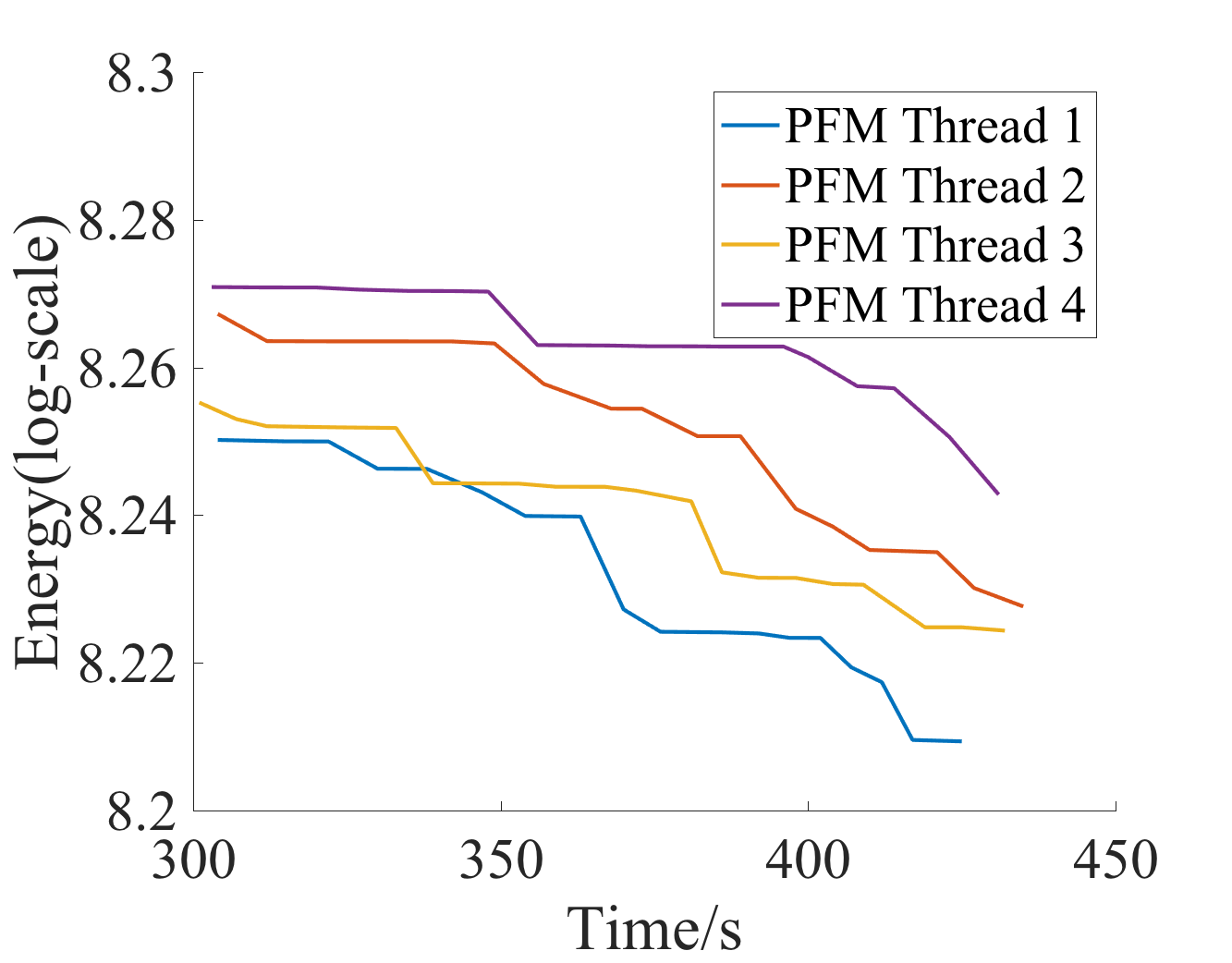}
  \end{subfigure}  
  \begin{subfigure}[b]{0.49\columnwidth}
    \centering
    \includegraphics[width=\columnwidth]{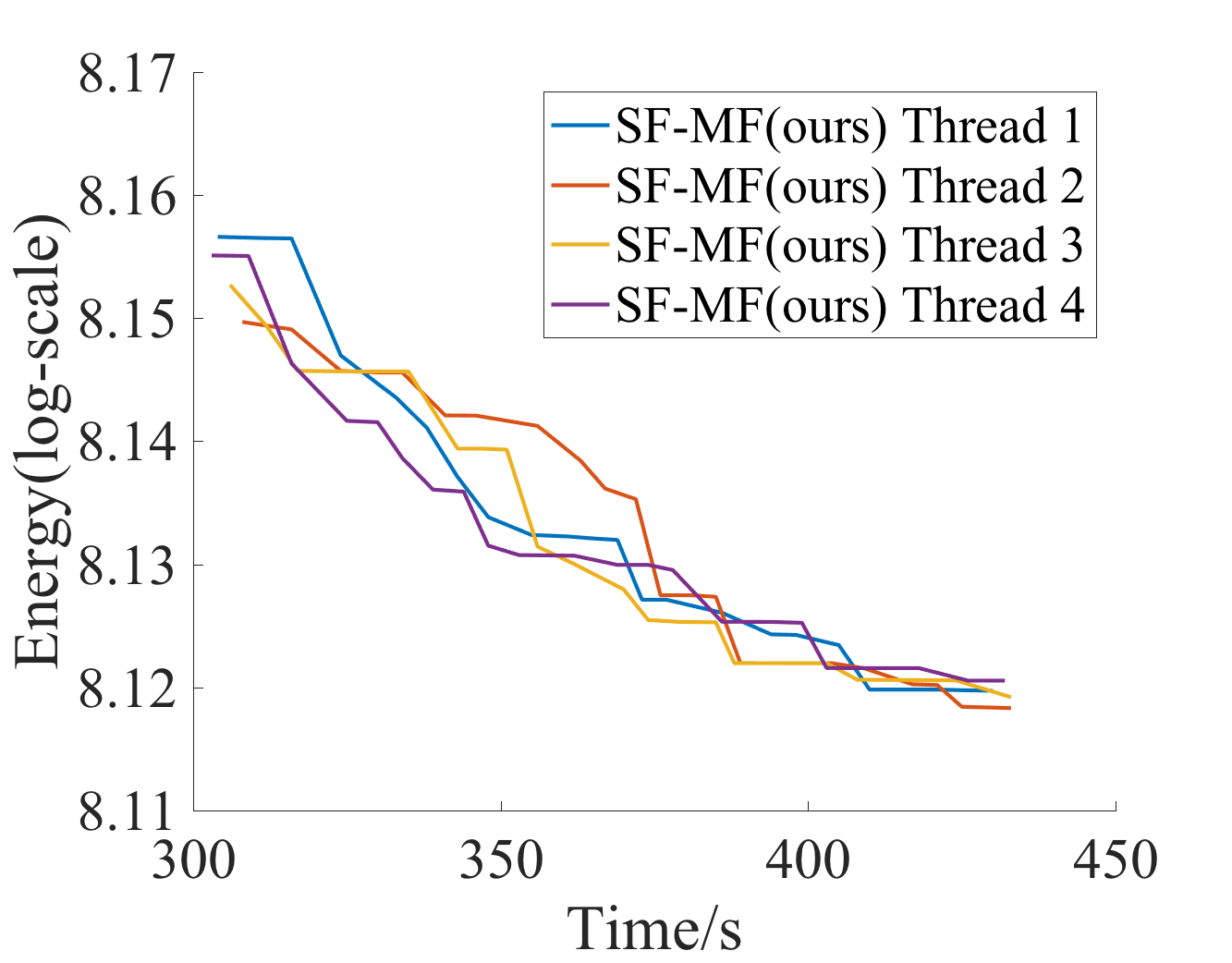}
  \end{subfigure}
  \caption{Energy plots per thread for the optical flow problem. Left: Parallel Fusion  
    Move (PFM). Right: SF-MF(ours).}
  \label{fig:optical_flow_by_threads}  
\end{figure}

% 
% This is because some solution proposals are more effective than others,
% so once a thread grabs an effective solution proposal, it find a lower
% energy quickly. Since there is no solution sharing in PFM model, other
% threads cannot share this lower energy state, and keeps working on its
% own state.
%
%On the other hand, SF-MF
%allows solution sharing, so once a thread grabs an effective solution
%proposal and moves to a lower energy state, other threads can share
%information about this lower energy state. In this manner, all threads
%contribute to further decrease of this low energy state. To further
%demonstrate what is happening here,
To further investigate the effectiveness of solution sharing,
Figure~\ref{fig:optical_flow_by_threads} shows the energy plots of PFM
and SF-MF per thread. As evident from the plot, in PFM, threads need to
keep working independently at higher energy states.
%in PFM, one thread found a better solution than others, but other
%threads keep working independently at higher energy states.
SF-MF, on the other hand, exchanges solutions all the time, and every
thread is making an effective work in improving the solution.
%As we can see from the plots, in PFM model, one thread finds lower
%energy state faster than others, while other threads keep working at
%their own energy state. But in SF-MF model, all threads exchange
%information about the lowest energy state frequently and work on
%improving the lowest energy state together. Since the solution for
%optical flow can be locally improved by each thread, the final merging
%of PFM can effectively fuse good local results in different threads
%together and achieve a similar energy state with SF-MF model. But as
%SF-MF model shares information in the middle, a final merging becomes
%less necessary.
Another key finding from Fig.~\ref{fig:optical_flow_convergence} is that
SF is slower than SF-MF. Our analysis is that multi-way fusion is
inefficient in this problem setting, since solution proposals are
relatively independent and fusing the solution space would not gain much
benefit. It rather loses performance against QPBO due to the
overhead of TRW-S.

\begin{figure}[!ht]
\centering
\begin{subfigure}[b]{0.49\columnwidth}
\includegraphics[width=\columnwidth]{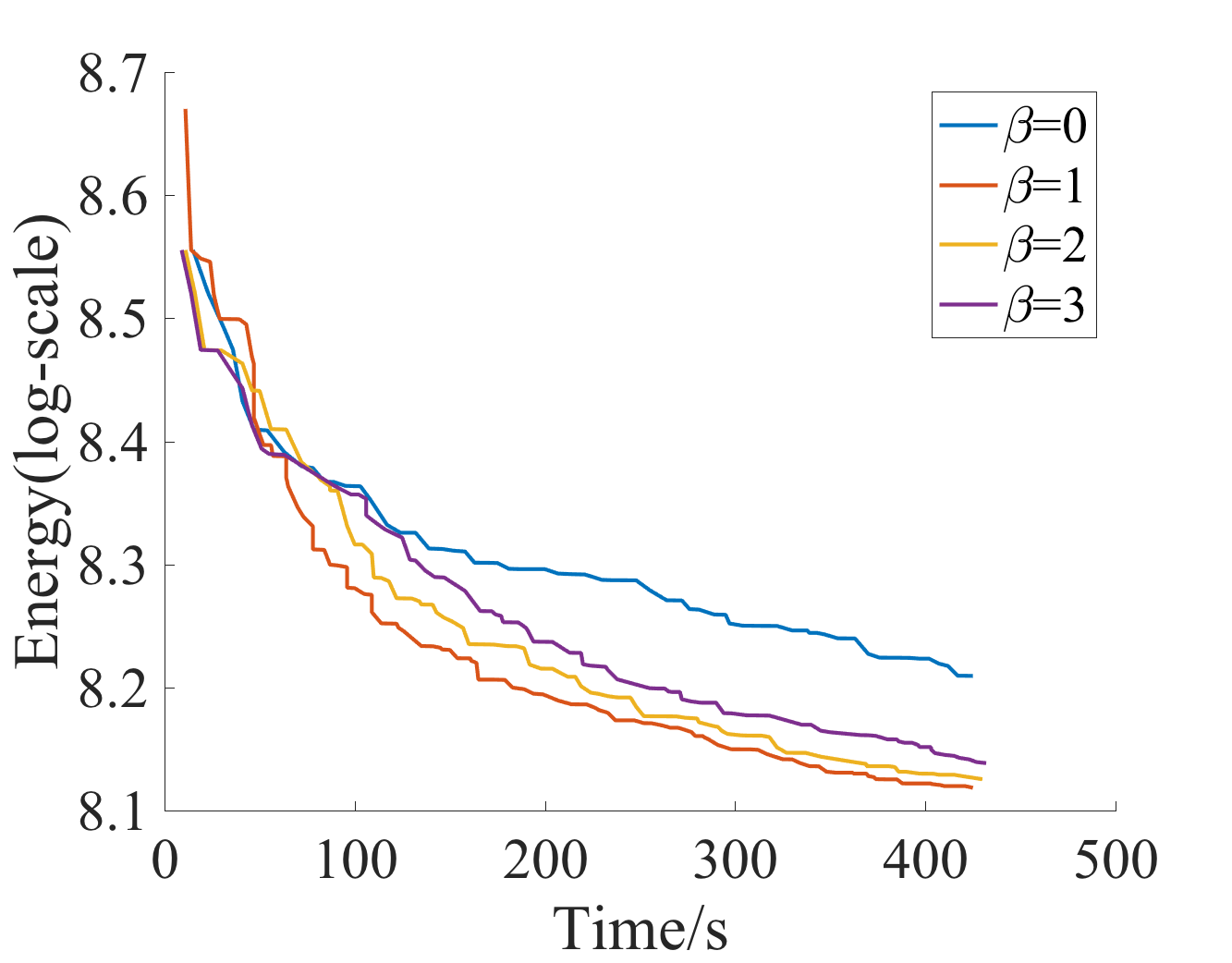}
\end{subfigure}
\begin{subfigure}[b]{0.49\columnwidth}
\includegraphics[width=\columnwidth]{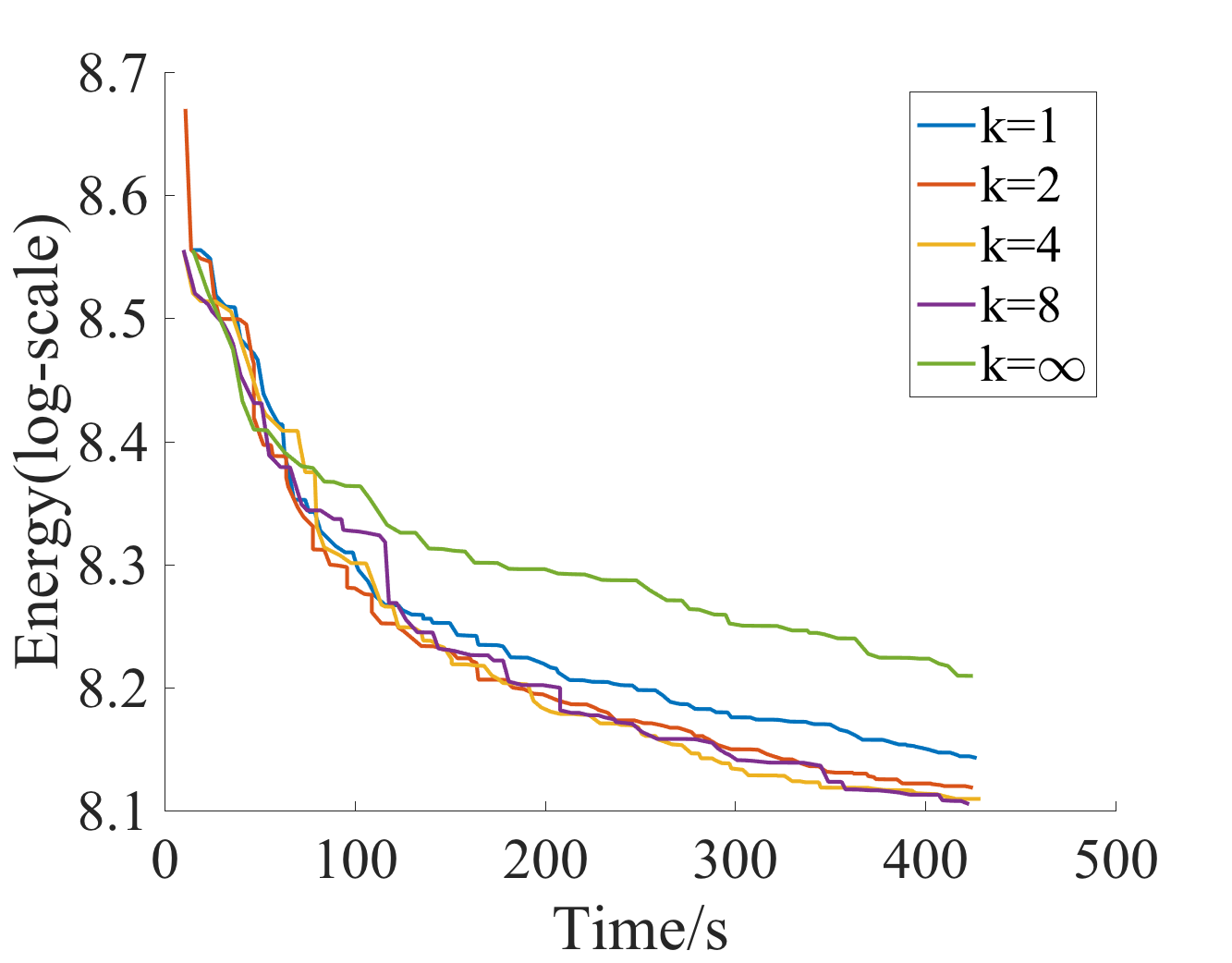}
\end{subfigure}
\caption{Energy
  plots for optical flow under different configurations. Left:
  varying $\beta$. Right: varying solution sharing
  frequencies. Solution sharing achieves better convergence, but
  sharing too many solutions (larger $\beta$) or sharing solutions too
  frequently (less $k$) slows down the convergence, as it reduces the
  time for exploration.} \label{fig:opticalflow_configuration}
\end{figure}

There are two factors influencing solution sharing: 1) \textit{the
number of solutions to share} and 2) \textit{the frequency of solution
sharing}. Both factors are controlled by $\beta$. As mentioned in
Section~\ref{section:optical_flow}, we have used $\beta = 1$ (i.e., share
solutions) once in every five iterations.  To further understand the
effects of solution sharing, we conducted two more experiments. First,
we set $\beta$ to 0, 1, 2, or 3 in every five iterations,
% (when $\beta = 0$, it is the same with PFM without final merging).
while keeping all other parameters the same (See
Fig.~\ref{fig:opticalflow_configuration}(left)).  Second, we change the number of
iterations $k$ between the two consecutive solution sharing iterations
(See Fig.~\ref{fig:opticalflow_configuration}(right)).  The first experiment
revealed that the solution sharing makes convergence faster regardless of
$\beta$.  However, too much solution sharing slows down the convergence,
and $\beta=1$ is the sweet spot for this problem.
%
% , energy generally
% decreases faster with solution sharing. But since we need to perform one
% fusion for sharing each solution, sharing more solution decreases
% efficiency in this problem setting. While sharing multiple solutions
% might be useful in some other problem setting as it means each thread
% can get more global information early.
%
The second experiment has shown that too frequent solution sharing
harms the convergence, simply because threads have less time generating more
proposals and exploring the solution space.
% From this figure, we can see that
% although solution sharing generally speeds up the optimization process,
% sharing solution too frequently is not a good practice. This is because
% when we share solution frequently, we have less time for generating and
%fusing new proposals.
Optimal parameter setting depends on each problem setting.
%A good choice of the solution sharing frequency depends on specific
%problem setting.

\mysubsubsection{Layered depthmap estimation}

\begin{figure}[!h]
  \centering
  \includegraphics[width=0.8\columnwidth]{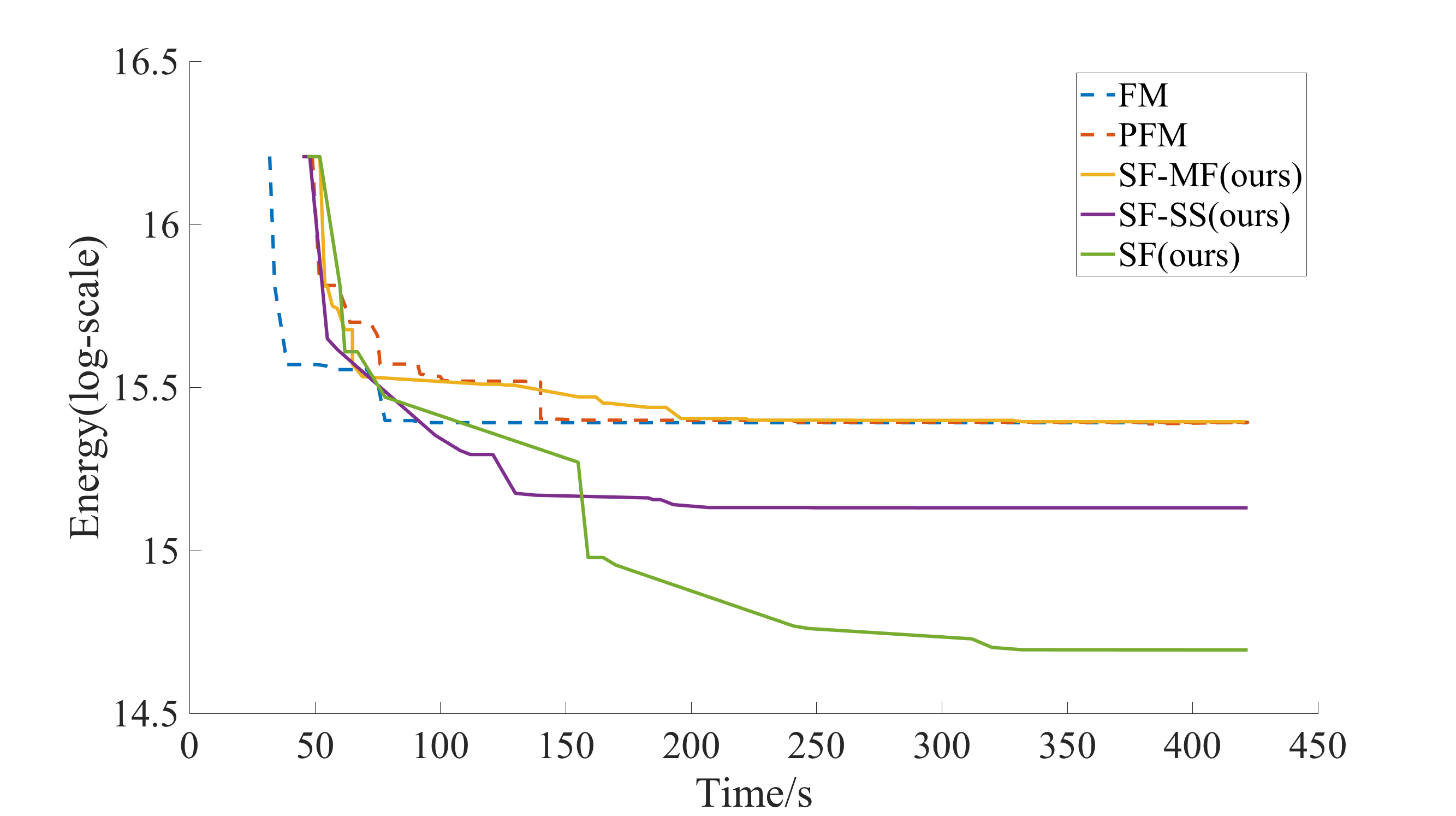}
  \caption{Energy plots for the layered depthmap estimation
    problem. Both the multi-way fusion and the solution sharing are important
    for this challenging problem.}\label{fig:layered_depthmap_convergence}
\end{figure}

\noindent We have used ``ours\_1'' data in~\cite{layered_depthmap} for
the experiments. Figure.~\ref{fig:layered_depthmap_convergence} shows that
Fusion Move, Parallel Fusion Move and SF-MF all got stuck in local
minima, which is due to the lack of multi-way fusion.  Layered depthmap
estimation is a challenging problem with very large solution space. The
binary fusion of solution proposals is too restrictive to make any
improvements.  This coincides with the observation in
\cite{layered_depthmap} that binary fusion of proposal solutions is not
as powerful as their subspace fusion which is a special form of
multi-way fusion here. Lastly, solution sharing also plays an important
role for this challenging problem, as SF performs much better than
SF-SS.

\begin{figure}[!h]
  \centering
  \begin{subfigure}[b]{0.49\columnwidth}
    \centering
    \includegraphics[width=\columnwidth]{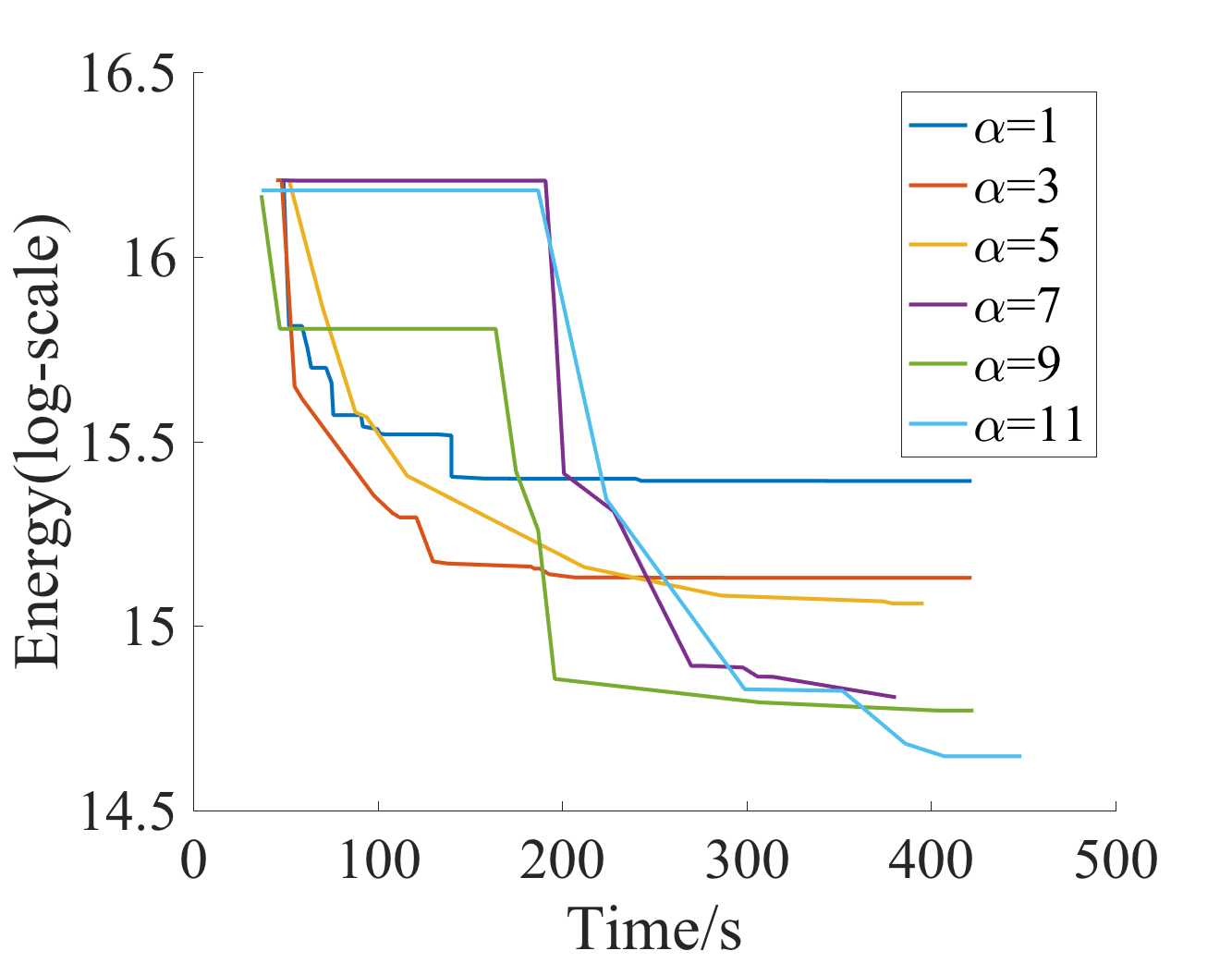}
  \end{subfigure}  
  \begin{subfigure}[b]{0.49\columnwidth}
    \centering
    \includegraphics[width=\columnwidth]{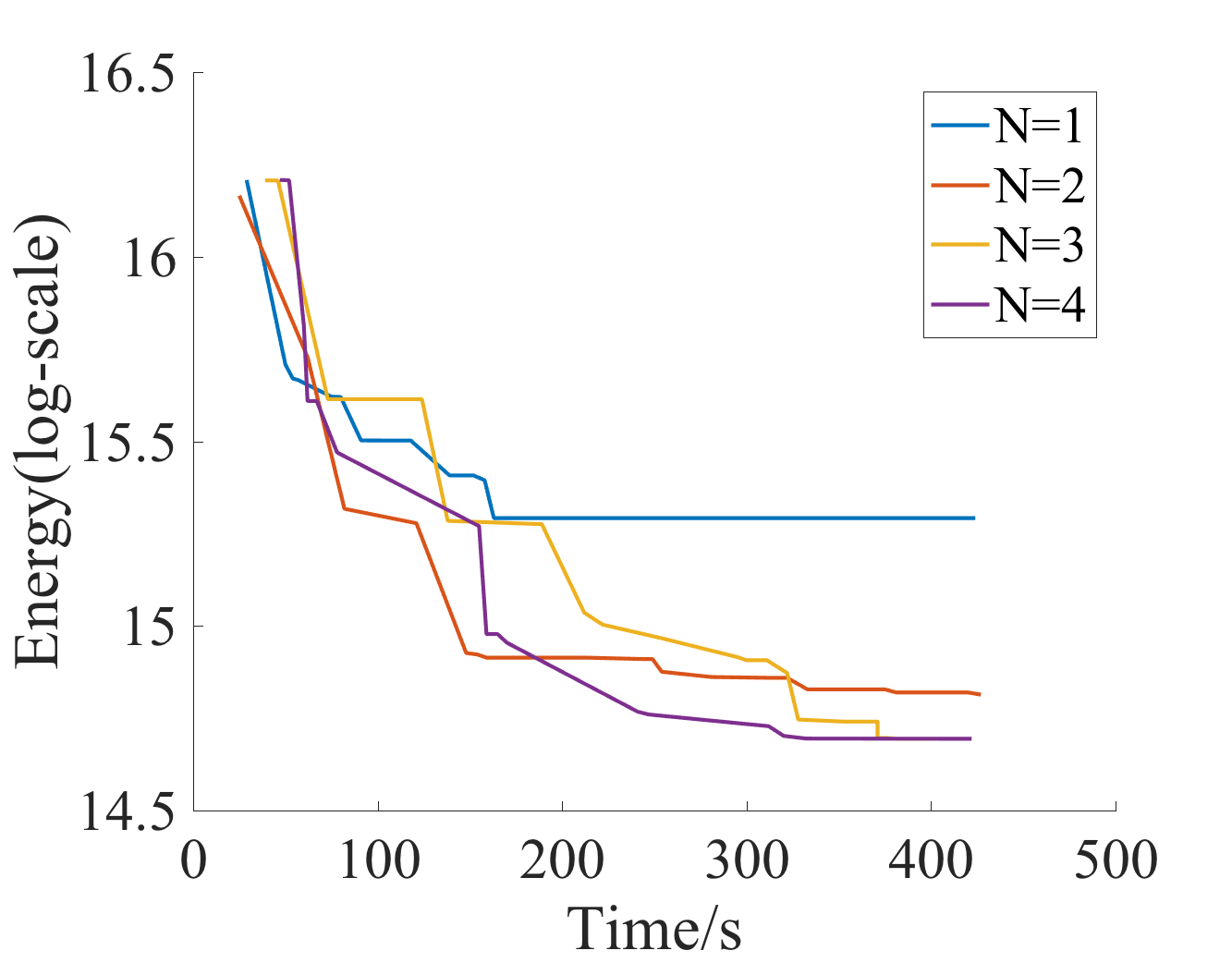}
  \end{subfigure}
  \caption{Left: Energy plots for layered depthmap estimation with varying
    $\alpha$. Right: Energy plots for layered depthmap estimation with varying number of threads N.}
  \label{fig:layered_depthmap_configuration}
\end{figure}

%
% From the plot, we can see that, Fusion Move, Parallel Fusion Move, and
%SF-MF all stalk at a high energy state.
To further study the effects of multi-way fusion, we have varied the
value of $\alpha$ which controls the number of solution proposals to be
fused in SF-SS model (See
Fig.~\ref{fig:layered_depthmap_configuration}(left)). Note that we have used SF-SS
instead of SF to disable solution sharing and better observe the effects
of multi-way fusion.
% SF-SS model (disable solution sharing to
%better observe the effect of multi-way fusion) while keeping other
%parameters the same and plot the energy minimization process in figure
It is interesting to see that more multi-way fusion takes longer to
converge, but finds a lower energy state at the end.

Finally, we have examined the role of multi-threading by varying the
number of threads N in our most general model SF (See
Fig.~\ref{fig:layered_depthmap_configuration}(right))~\footnote{While keeping other
parameters the same, we have to change $\beta$ with $N$ because of the
constraint $\beta \leq N-1$. We have always used $\beta = N-1$ in this
experiment.}. More threads lead to faster convergence as expected,
although the rate of speed-up is not proportional to the number of
threads due to the randomness in the proposal generation scheme.
%The plot shows that SF finds low energy faster with more
%threads.
% In Fusion Move like methods, parallelism speeds up convergence
% by exploring solution proposals faster. Due to the inherited randomness
% in the proposal generation scheme, the speed up is often not
% proportional to the number of threads.
%We can see that as the number of
%ways to fuse becomes larger, each fusion step takes longer time, but the
%chance of finding a lower energy state increases.

%As shown in the above problem settings, the solution sharing and
%multi-way fusion enabled by our uniform framework play a key role for
%improving performance in different problem settings.

\section{Conclusion and future directions}
We have proposed a novel MRF inference framework, Swarm Fusion, in
parallel computing environments. The framework is general and makes
popular inference techniques such as Alpha Expansion, Fusion Move,
Parallel Alpha Expansion, and Hierarchical Fusion, its special cases. Our
experiments have revealed that the framework exploits parallel
computational resources and achieves faster convergence, especially for
challenging problems.  Our first future work is to conduct experiments
on cloud computing environments, in particular, the MapReduce
programming model, where the roles of mappers and reducers exactly
correspond to the processes of parallel multi-way fusion and solution
sharing, respectively.  Another future work is the automatic
configuration of the Swarm Fusion architecture.  Our experiments have
shown that optimal architectures are different for different problems.
An interesting direction is to adaptively change its architecture during
the computation, for example, switching to simple parallel
alpha-expansion for easy problems, or increasing the rate of solution
exchanges when solutions vary significantly across threads.
Parallel MRF inference has been a relatively under-explored topic in
Computer Vision.
%
%Our idea is very simple, where adapting the Swarm Fusion
%framework require minimal coding.
%
The proposed Swarm Fusion framework can be intergrated into existing
algorithms with minimal coding.  We believe that this paper would
immediately benefit tens of thousands of Computer Vision researchers
or engineers in the world, who currently solve MRF problems. We will
share our source code with the community.

\section{Acknowledgement}
This research was supported by National Science Foundation
under grant IIS 1540012 and Google Faculty Research
Award. We thank Nvidia for a generous GPU donation.

\clearpage
\bibliographystyle{splncs}
\bibliography{eccv16bib}
\end{document}